\documentclass[conference]{IEEEtran}
\IEEEoverridecommandlockouts
\usepackage{cite}
\usepackage{amsmath,amssymb,amsfonts}
\usepackage{graphicx}
\usepackage{textcomp}
\usepackage{xcolor}
\usepackage{booktabs, adjustbox}
\usepackage{comment}
\usepackage{url}
\usepackage{algorithm}

\usepackage[noend]{algpseudocode}
\def\BibTeX{{\rm B\kern-.05em{\sc i\kern-.025em b}\kern-.08em
    T\kern-.1667em\lower.7ex\hbox{E}\kern-.125emX}}
\begin{document}

\title{SUTA-LM: Bridging Test-Time Adaptation and Language Model Rescoring for Robust ASR}

\author{
\IEEEauthorblockN{Wei-Ping Huang$^*$\thanks{$^*$Equal contribution}, Guan-Ting Lin$^*$, Hung-yi Lee}
\IEEEauthorblockA{\textit{Graduate Institute of Communication Engineering} \\
\textit{National Taiwan University}\\
Taipei, Taiwan \\
thomas1232121@gmail.com, \{f10942104, hungyilee\}@ntu.edu.tw}
}

\maketitle

\begin{abstract}
Despite progress in end-to-end ASR, real-world domain mismatches still cause performance drops, which Test-Time Adaptation (TTA) aims to mitigate by adjusting models during inference. Recent work explores combining TTA with external language models, using techniques like beam search rescoring or generative error correction. In this work, we identify a previously overlooked challenge: TTA can interfere with language model rescoring, revealing the nontrivial nature of effectively combining the two methods. Based on this insight, we propose SUTA-LM, a simple yet effective extension of SUTA, an entropy-minimization-based TTA approach, with language model rescoring. SUTA-LM first applies a controlled adaptation process guided by an auto-step selection mechanism leveraging both acoustic and linguistic information, followed by language model rescoring to refine the outputs. Experiments on 18 diverse ASR datasets show that SUTA-LM achieves robust results across a wide range of domains.\footnote{The source code is available at \url{https://github.com/hhhaaahhhaa/ASR-TTA}}
\end{abstract}


\section{Introduction}
Automatic Speech Recognition (ASR) has made significant progress with end-to-end deep learning models. However, real-world deployment remains challenging due to domain shifts, such as acoustic and linguistic variability, which can substantially degrade performance on out-of-domain data.
Test-Time Adaptation (TTA) is an appealing strategy to improve model robustness under distribution shifts. It adapts the source model at inference time using only the test data, without requiring access to the original training distribution. TTA is typically driven by unsupervised objectives, such as entropy minimization~\cite{tent, eata, stable-tta}, or pseudo-labeling~\cite{cotta, cpl}. While originally developed for computer vision tasks like image classification, TTA has recently gained momentum in ASR~\cite{suta, sgem, awmc, lin-etal-2024-continual, litta}, demonstrating promising results.

In parallel, language model (LM) rescoring~\cite{hannun2014deepspeechscalingendtoend} is a widely used technique in ASR that enhances linguistic accuracy by incorporating external LMs during beam search decoding. Despite the widespread adoption of both techniques, prior work has treated TTA and LM rescoring as independent strategies. To the best of our knowledge, there has been little to no attempt to systematically combine acoustic adaptation through TTA with linguistic adaptation through LM rescoring in a unified framework.

Among existing work, SGEM~\cite{sgem} comes closest to connecting these two components. It uses beam hypotheses generated with an external LM to calibrate ASR logits before adaptation. However, the adaptation itself is still governed solely by an entropy-based TTA objective, and LM-based decoding is not used after adaptation. As a result, SGEM does not fully realize the potential of jointly leveraging acoustic and linguistic adaptation, and its approach introduces additional computational complexity.

\begin{figure}
\centering
\includegraphics[width=1.0\linewidth]{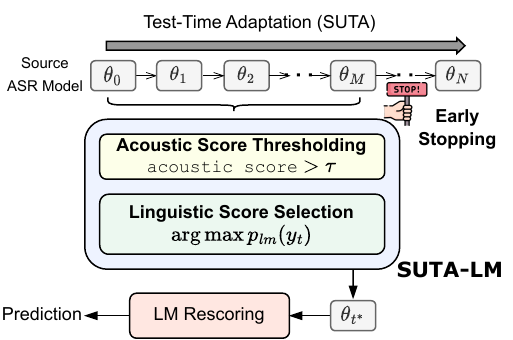}
\caption{An overview of the proposed SUTA-LM method, which integrates acoustic score thresholding, linguistic score selection, and early stopping to automatically select the best adapted model $\theta_{t^*}$ for LM rescoring.}
\label{fig:overview}
\end{figure}

In this work, we aim to investigate the strategy to integrate TTA with LM rescoring. Our goal is to adapt the acoustic model using TTA and then apply LM rescoring during decoding to inject linguistic knowledge. Beginning with the preliminary study (Section~\ref{sec:prelim}) on this simple sequential baseline, we reveal that the number of adaptation steps used in TTA plays a pivotal role. Inappropriate adaptation—whether too conservative (too few steps) or too aggressive (too many steps)—can degrade the effectiveness of LM rescoring, resulting in suboptimal performance. Building on this observation, we propose SUTA-LM, a simple yet effective approach extending the previous work, SUTA\cite{suta}, with LM rescoring.

Fig.~\ref{fig:overview} provides an overview of the proposed method. SUTA-LM incorporates a carefully designed auto-step selection mechanism, which dynamically adjusts the number of adaptation steps for each sample. Auto-step selection consists of two components: \textbf{(1) Acoustic Score Thresholding}, which determines if the adaptation process has not yet reached a sufficient level, and \textbf{(2) Linguistic Score Selection}, which selects the optimal adaptation step based on LM scores. Our method leverages both acoustic and linguistic information for the final decision. Furthermore, the auto-step selection mechanism can be implemented in an online fashion with an early stopping strategy. If the decision remains unchanged over multiple iterations, the adaptation process is terminated early, which significantly improves efficiency.

Our contributions can be summarized as follows:
\begin{itemize}
    \item We identify that inappropriate adaptation processes can negatively impact the effectiveness of LM rescoring, highlighting the need for careful control of the adaptation process.

    \item We introduce a robust auto-step selection mechanism that effectively combines LM rescoring and TTA, ensuring consistent ASR performance across diverse scenarios.

    \item We provide a comprehensive comparison between beam search decoding with LM rescoring and TTA for ASR, addressing practical scenarios that have been largely overlooked in previous work.
\end{itemize}

\section{Related Work}

Test-time adaptation (TTA) in ASR aims to improve performance on unseen domains without requiring labeled data. SUTA~\cite{suta} is an entropy-minimization approach that iteratively adapts the model by minimizing a combined objective of output entropy and class confusion over a fixed number of steps, reducing uncertainty and sharpening decision boundaries. Liu \textit{et al.}\cite{cea} improve robustness through confidence-based adaptation and short-term consistency regularization. AWMC\cite{awmc} adopts a teacher-student framework for continual TTA, while DSUTA~\cite{lin-etal-2024-continual} stabilizes long-term adaptation via a hybrid fast-slow model. However, these methods focus solely on acoustic adaptation, often overlooking external linguistic cues and relying on greedy decoding.



Recognizing this gap, recent efforts have explored integrating LMs into TTA. SGEM~\cite{sgem} combines sequential-level generalized entropy minimization with beam search logit calibration, while LI-TTA~\cite{litta} leverages pseudo labels generated by LLMs using a generative error correction (GEC) framework~\cite{10389673, hu-etal-2024-listen}. 
While these approaches advance LM integration, they incur \textit{high computational costs}, and \textit{LM-based decoding is not utilized after adaptation}. Whether a simpler and more efficient method for integrating TTA with LM rescoring can achieve comparable performance remains an open question.

In this work, we address these issues by providing a comprehensive comparison of LM rescoring and TTA in ASR, revealing that a carefully tuned combination of these methods can lead to improved performance. By introducing minimal modifications to the sequential baseline, our method maintains high efficiency while effectively integrating LMs into TTA.

\section{Preliminary Study}
\label{sec:prelim}
The goal of this preliminary study is to assess whether sequentially applying SUTA and LM rescoring can lead to superior ASR performance. Specifically, we aim to compare the performance of the sequential baseline, \textbf{SUTA+Rescoring}, with \textbf{Rescoring} alone. 
To formalize SUTA+Rescoring, denote the source ASR model as $\theta$. For each incoming test utterance $x$, SUTA is applied to adapt $\theta$ over $N$ iterative steps, resulting in a sequence of adapted models: $\theta_0 = \theta, \theta_1, \ldots, \theta_N$. Beam search decoding with LM rescoring is then performed on the \textbf{final} output logits $\theta_N(x) \in \mathbb{R}^{L\times C}$, where $L$ is the number of frames and $C$ is the vocabulary size.

This initial analysis focuses on understanding how the test-time adaptation process influences the effectiveness of LM rescoring and whether SUTA+Rescoring can achieve robust results across diverse test scenarios.

\subsection{Experimental Setup}
\label{sec:prelim-implement}
For SUTA, we follow the official implementation\footnote{https://github.com/DanielLin94144/Test-time-adaptation-ASR-SUTA}, where an additional reweighting trick is applied to the minimum class confusion loss. The source ASR model is the wav2vec 2.0-base model fine-tuned on
Librispeech 960 hours\footnote{https://huggingface.co/facebook/wav2vec2-base-960h} and the hyperparameters are set to the same as the original paper. For the LM rescoring, we use the external 4-gram language
model\footnote{https://huggingface.co/patrickvonplaten/wav2vec2-base-960h-4-gram}, and the beam search decoding implementation from pyctcdecode, which mainly follows the shallow fusion approach as in \cite{hannun2014deepspeechscalingendtoend}: \begin{equation*}
    \hat{y} = \arg\max_y\, \log p_{\theta}(y|x) + \alpha\log p_{lm}(y) + \beta\, \text{length}(y)
\end{equation*}
where $\alpha$ and $\beta$ are tunable parameters, $p_{\theta}$ is the probability from the ASR model, $p_{lm}$ is the probability from the 4-gram model, and $x$ is the input utterance.
In this work, we fix $(\alpha, \beta)=(0.5, 0)$ in all the experiments for simplicity.

For this preliminary analysis, we selected datasets that capture a range of typical domain shifts, including synthetic 10dB Gaussian noise (LS-C~\cite{lin-etal-2024-continual}), real-world environments (TED-LIUM3~\cite{ted3}), and Spanish accent (L2Arctic~\cite{l2arctic}). This selection allows us to highlight the main observations without overwhelming the analysis with too many cases. The full benchmark will be introduced in Section~\ref{sec:benchmark}.

\subsection{Key Observations}
\label{sec:key}
\noindent\textbf{(1) SUTA may undermine the effectiveness of LM rescoring.} Table~\ref{tab:prelim1} presents the word error rates (WER) for \textbf{SUTA+Rescoring} and \textbf{Rescoring} across different datasets, along with the WERs of the source model and SUTA for reference. While \textbf{SUTA+Rescoring} achieves a notable improvement on the Gaussian noise (27.3 WER vs. 38.9 for \textbf{Rescoring}), the trend reverses on the other datasets. On TED-LIUM3 and the Spanish accent, \textbf{SUTA+Rescoring} yields higher WERs (11.4 and 16.5) than \textbf{Rescoring} (11.2 and 15.4), indicating that the straightforward combination of the two methods can degrade performance. Neither method is consistently superior across different domains, therefore requiring a more general solution.

\begin{table}
\caption{WER (\%) comparison over different TTA methods. Reported WER is averaged over 3 runs.}
\label{tab:prelim1}
    \centering
        \begin{tabular}{lccc}
            \toprule
            \textbf{Method} & \textbf{Gaussian} & \textbf{TED-LIUM3} & \textbf{Spanish} \\
            \midrule
            \textbf{Source model} & 45.1 & 13.0 & 21.6 \\
            \textbf{SUTA} & 28.1 & 11.7 & 17.7 \\
            \midrule
            \textbf{Rescoring} & 38.9 & \textbf{11.2} & \textbf{15.4} \\
            \textbf{SUTA+Rescoring} & \textbf{27.3} & 11.4 & 16.5 \\
            \midrule
        \end{tabular}
\end{table}

Interestingly, \textbf{Rescoring} can be viewed as a special case of \textbf{SUTA+Rescoring} with zero adaptation steps ($N = 0$), while the original SUTA fixes $N = 10$. This naturally raises the question: \textit{\textbf{how does varying the number of adaptation steps influence overall performance?}} To investigate this, we vary the number of adaptation steps from 0 to 20, effectively exploring a continuum from pure rescoring to aggressive adaptation with rescoring. This leads to our second key observation.

\begin{figure}[t]
\centering
\includegraphics[width=1.0\linewidth]{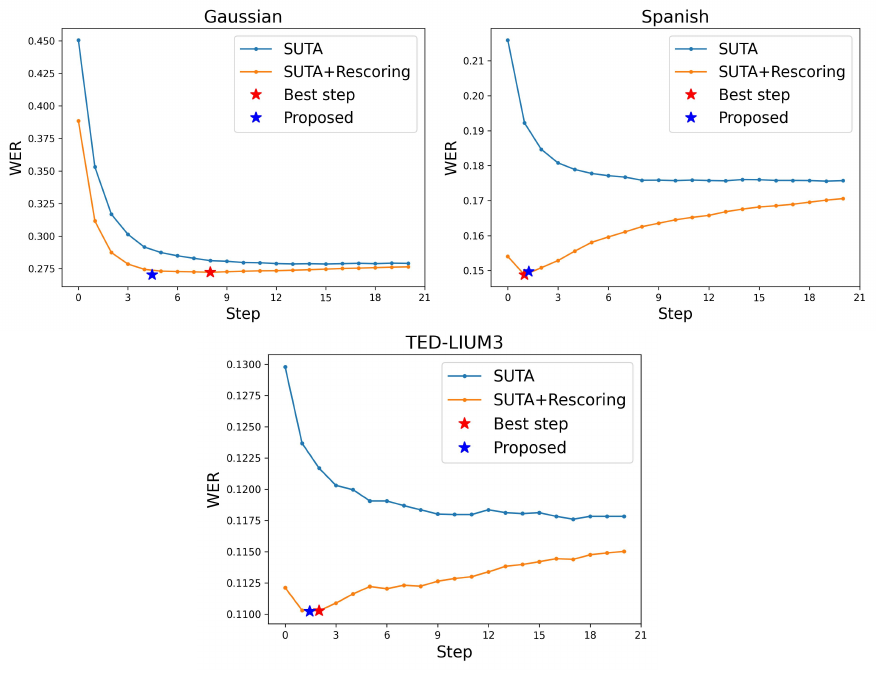}
\caption{WER(\%) of SUTA and SUTA+Rescoring over different numbers of adaptation steps on Gaussian noise, TED-LIUM3, and the Spanish accent. The best number of steps is marked by a red star. Our proposed method is marked by a blue star for reference.}
\label{fig:prelim2}
\end{figure}

\noindent\textbf{(2) The optimal number of adaptation steps varies significantly across different datasets.} Fig.~\ref{fig:prelim2} shows the WER over different numbers of adaptation steps on Gaussian noise, TED-LIUM3, and the Spanish accent. The optimal number of steps, marked by a red star for each dataset, ranges widely: Gaussian noise requires 8 steps, the Spanish accent only 1 step, and TED-LIUM3 just 2 steps. This indicates that the optimal level of adaptation depends on the characteristics of the input data. An inappropriate adaptation strategy—whether too conservative (too few steps) or too aggressive (too many steps)—can lead to suboptimal performance. Moreover, \textbf{Rescoring} (\textbf{SUTA+Rescoring} with $N=0$) is suboptimal in all three settings, suggesting that a carefully tuned combination of model adaptation and rescoring has the potential to deliver better results.

Additionally, Fig.~\ref{fig:prelim2} shows the WER of SUTA across varying numbers of adaptation steps. As the number of steps increases, the performance of SUTA with greedy decoding remains relatively stable. However, the WER of \textbf{SUTA+Rescoring} begins to degrade noticeably. This discrepancy suggests that even if adaptation appears effective or has minimal impact under greedy decoding, it may adversely affect the outcome after LM rescoring, which highlights the nontrivial interaction between the adaptation process and LM rescoring.

\section{Method}
Based on the findings in Section~\ref{sec:key}, our goal is to design an \textit{\textbf{auto-step selection}} mechanism that dynamically determines the optimal number of adaptation steps in SUTA for each test sample, aiming to maximize performance after LM rescoring.

The following subsections detail our proposed method. Section~\ref{sec:ass} outlines the overall auto-step selection framework. Sections~\ref{sec:am} and~\ref{sec:lm} then introduce its two core components: \textbf{Acoustic Score Thresholding} and \textbf{Linguistic Score Selection}, respectively. Finally, Section~\ref{sec:early-stop} presents an early stopping strategy that further reduces unnecessary adaptation steps, improving the method’s overall efficiency. 

\subsection{Auto-step Selection}
\label{sec:ass}
Rather than always selecting the final adapted model $\theta_N$ as in SUTA+Rescoring, our proposed auto-step selection algorithm dynamically selects an intermediate model $\theta_t$ for some $0 \leq t \leq N$, based on the input $x$ and the entire adaptation trajectory:
\begin{equation}
\label{eq:a}
    t = f(x, \theta_0, \theta_1, \cdots, \theta_N),
\end{equation}
with the goal of minimizing the WER of the transcription decoded from $\theta_t(x)$ with beam search.

The selection algorithm $f$ is constructed using two components: \textbf{Acoustic Score Thresholding} and \textbf{Linguistic Score Selection}, which jointly consider acoustic and linguistic signals to determine the optimal step.

\subsection{Acoustic Score Thresholding}
\label{sec:am}

We first eliminate adaptation steps that are likely to be unreliable based on the model's confidence. Let $c^t_l$ denote the confidence of the $l$-th frame in the output logits $\theta_t(x)$, defined as the probability of the predicted class for that frame. The acoustic score for step $t$ is defined as the average log confidence over all frames:
\begin{equation}
    S_t=\frac{1}{L}\sum_{l=1}^L \log c^t_l.
\end{equation}
We retain only the steps with high acoustic score:
\begin{equation}
    \mathcal{T} = \left\{S_t \geq \tau | 0\leq t \leq N \right\} ,
\end{equation}
where $\tau$ is a predefined threshold. The subset $\mathcal{T}$ represents all steps where the model exhibits sufficient confidence.

\subsection{Linguistic Score Selection}
\label{sec:lm}
Since all steps in $\mathcal{T}$ are deemed acoustically plausible, we further distinguish among them using a linguistic criterion. Denote the greedy decoded transcription from the output logits $\theta_t(x)$ as $y_t$, we select the step $t$ where $y_t$ achieves the highest probability under the external language model:
\begin{equation}
    t^* = \arg\max_{t\in\mathcal{T}}\, p_{lm}(y_t).
\end{equation}
If multiple steps achieve the same maximum probability, we choose the one with the smallest index. In case that $\mathcal{T} = \emptyset$, we default to selecting the final step $t^*=N$. This preference for earlier steps is motivated by our preliminary analysis, which suggests that over-adaptation can sometimes harm performance after LM rescoring.

Together, the two-stage selection process takes into account both acoustic reliability and linguistic plausibility, increasing the likelihood that applying LM rescoring to $\theta_{t^*}(x)$ produces a high-quality transcription.

\subsection{Early Stopping}
\label{sec:early-stop}

The auto-step selection mechanism can be implemented in an online fashion with an early stopping strategy, which improves computational efficiency by potentially terminating the adaptation process before all steps are completed. Specifically, we continue adapting the model step-by-step for steps $t \notin \mathcal{T}$, and track only the steps $t \in \mathcal{T}$, where the acoustic score exceeds the predefined threshold. For these valid steps, we monitor the linguistic score $p_{lm}(y_t)$, and the adaptation process is terminated early once the best observed linguistic score does not improve for $P$ consecutive valid steps. At that point, the step corresponding to the highest linguistic score encountered so far is selected. If no steps satisfy the acoustic score thresholding criterion, we default to selecting the final step $t^*=N$. While the original formulation~\eqref{eq:a} requires executing all $N$ adaptation steps, using early stopping can halt the process earlier, reducing computation and improving overall efficiency. 

\section{Experiments}
\label{sec:exp}

\subsection{Datasets}
\label{sec:benchmark}
\noindent\textbf{Corrupted Librispeech (LS-C)~\cite{lin-etal-2024-continual}}: The dataset is constructed by adding noises from MS-SNSD~\cite{ms-snsd} into Librispeech test set~\cite{librispeech}. The noises include air conditioner (AC), airport announcement (AA), babble (BA), copy machine
(CM), munching (MU), neighbors (NB), shutting
door (SD), typing (TP), vacuum cleaner (VC), and Gaussian noise (GS), resulting in 10 different noises in total. The
Signal-to-Noise Ratio (SNR) is set to 10 dB. 
\\
\noindent\textbf{L2Arctic~\cite{l2arctic}}: A non-native English speech corpus consisting of utterances from second language (L2) learners originating from 6 countries with different first languages (L1): Arabic, Mandarin, Hindi, Korean, Spanish, and Vietnamese.
\\
\noindent\textbf{TED-LIUM3 (TD)~\cite{ted3}}: English speech extracted from TED conference talks and the corresponding transcriptions. The recordings are generally clean but may contain mild reverberation due to the recording environment.
\\
\noindent\textbf{CHiME-3 (CH)~\cite{chime}}: A noisy version
of WSJ corpus mixed with recordings made in 4 noisy environments (Cafe, Bus, Street, Pedestrian Area).

Our selection of benchmarks is designed to cover a diverse range of domains, including synthetic noise, accented speech, reverberant conditions, and real-world background interference, offering a comprehensive evaluation of ASR robustness under domain shift.

\subsection{Baselines}
We compare the proposed SUTA-LM with four baselines: \textbf{SUTA}, \textbf{Rescoring}, \textbf{SUTA+Rescoring}, \textbf{SGEM}. We report the results with maximal adaptation steps $N=20$.

\subsection{Implementation Details}
\label{sec:implement}

The implementation of SUTA, Rescoring, and SUTA+Rescoring follows Section~\ref{sec:prelim-implement}.
For SGEM, we follow the official implementation\footnote{https://github.com/drumpt/SGEM}. For the proposed SUTA-LM, the default threshold is $\tau=-0.05$, and the patience $P$ is set to 3.
Following~\cite{lin-etal-2024-continual}, we exclude samples with raw lengths longer than
20 seconds in all experiments, which accounts for less than 1\% of the data. All experiments were conducted on an Nvidia GeForce RTX 3080Ti GPU.

\subsection{Results}
\label{sec:results}

\begin{table*}[t]
  \centering
  \caption{WER (\%) of different TTA methods across 18 datasets. \textbf{Avg.} represents the averaged WER over all 18 datasets. Reported WER is averaged over 3 runs.}
      \begin{tabular}{lcccccccccc}
        \toprule
         \textbf{Method} & \textbf{AA} & \textbf{AC} & \textbf{BA} & \textbf{CM} & \textbf{GS} & \textbf{MU} & \textbf{NB} & \textbf{SD} & \textbf{TP} & \textbf{VC} \\
        \midrule
        \textbf{Source model}  & 21.6 & 14.2 & 38.1 & 28.5 & 45.1 & 30.6 & 81.7 & 13.6 & 16.6 & 29.5 \\
        \midrule
        \textbf{SUTA}  & 16.2 & 10.6 & 27.7 & 21.5 & 28.1 & 22.1 & 64.6 & 10.8 & 12.1 & 20.6 \\
        \textbf{Rescoring}  & 17.6 & 10.8 & 31.6 & 23.8 & 38.9 & 25.9 & 65.7 & 10.5 & 12.9 & 24.7 \\
        \textbf{SGEM} & 16.2 & 10.4 & 28 & 21.4 & 28.3 & 22.3 & 66.1 & 10.5 & 11.9 & 20.5 \\
        \textbf{SUTA+Rescoring} & 15.9 & 10.2 & 27.5 & 21.1 & 27.6 & 21.9 & 64.6 & 10.5 & 11.8 & 20.1 \\
        \textbf{SUTA-LM} & \textbf{15.2} & \textbf{9.6} & \textbf{26.9} & \textbf{20.2} & \textbf{27.0} & \textbf{21.2} & \textbf{63.8} & \textbf{9.7} & \textbf{11.1} & \textbf{19.4} \\
        \bottomrule
      \end{tabular}
      \\[\baselineskip]

      \begin{tabular}{lcccccc|c|c||c}
        \toprule
         \textbf{Method} & \textbf{Arabic} & \textbf{Chinese} & \textbf{Hindi} & \textbf{Korean} & \textbf{Spanish} & \textbf{Vietnamese} & \textbf{TD} & \textbf{CH} & \textbf{Avg.}\\
        \midrule
        \textbf{Source model}  & 21.4 & 27.5 & 17.1 & 17.3 & 21.6 & 34.8 & 13.0 & 30.0 & 27.9\\
        \midrule
        \textbf{SUTA}  & 17.5 & 22.1 & 13.1 & 14.1 & 17.7 & 30.2 & 11.7 & 23.5 & 21.3\\
        \textbf{Rescoring}  & \textbf{15.2} & 20.6 & 12.2 & 11.8 & 15.4 & 28.1 & 11.2 & 26.7 & 22.4\\
        \textbf{SGEM} & 17.3 & 22.2 & 12.9 & 13.9 & 17.5 & 30.1 & 11.6 & 23.4 & 21.4\\
        \textbf{SUTA+Rescoring} & 17.1 & 21.5 & 12.7 & 13.7 & 17.1 & 29.6 & 11.5 & 23.1 & 21.1\\
        \textbf{SUTA-LM} & \textbf{15.2} & \textbf{19.6} & \textbf{11.3} & \textbf{11.7} & \textbf{15.0} & \textbf{27.6} & \textbf{11.0} & \textbf{22.4} & \textbf{19.9}\\
        \bottomrule
      \end{tabular}
  \label{tab:main-exp}
\end{table*}

\textbf{SUTA-LM consistently outperforms across diverse domains.} 
Table~\ref{tab:main-exp} summarizes the results across all 18 datasets. The proposed SUTA-LM achieves an average WER of 19.9, outperforming SGEM, which has a WER of 21.4, while also being less complex. Compared to the individual components, SUTA (21.3 WER) and Rescoring (22.4 WER), SUTA-LM combines their strengths effectively. Although SUTA+Rescoring also integrates both approaches, it suffers from inconsistent performance due to the varying optimal number of adaptation steps across inputs, as mentioned in the preliminary study. In contrast, SUTA-LM addresses this issue through an auto-step selection mechanism and delivers better results (19.9 
 $<$ 21.1). This highlights the importance of dynamically selecting an appropriate number of adaptation steps to effectively leverage LM rescoring. Overall, SUTA-LM outperforms all other methods, across noisy (LS-C), accented (L2Arctic), and environment-shifted (TD, CH) domains, demonstrating strong generalization under diverse domain shifts.

\begin{figure}
\centering
\includegraphics[width=1.0\linewidth]{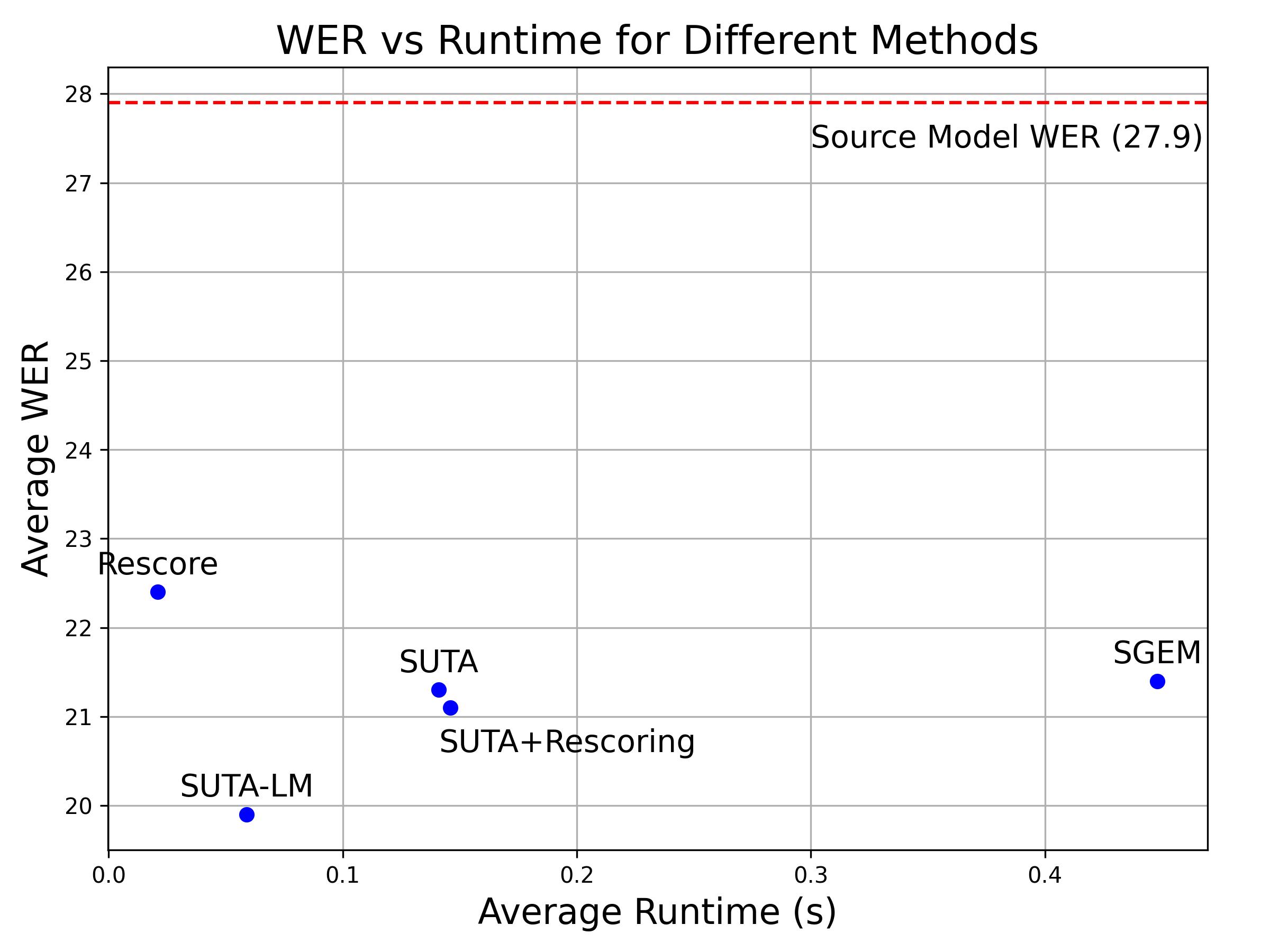}
\caption{Comparison of average WER and averaged runtime for a 1-second utterance for different methods. The dashed red line marks the WER of the unadapted source model. The results are averaged over 3 runs.}
\label{fig:runtime}
\end{figure}

\textbf{SUTA-LM demonstrates high efficiency.}
Fig.~\ref{fig:runtime} compares the average WER against average runtime (in seconds) for a 1-second utterance across different methods. SUTA-LM is faster than SUTA+Rescoring, which in turn is faster than SGEM. Specifically, SUTA-LM runs about 7× faster than SGEM and over 2× faster than SUTA+Rescoring, while still achieving the lowest WER among all methods. Although slightly slower than Rescore, it significantly outperforms it in accuracy. This high efficiency is due to SUTA-LM’s ability to reduce the number of adaptation steps per sample effectively. Overall, SUTA-LM offers a well-balanced trade-off between performance and efficiency.

\section{Discussion}
\subsection{SUTA-LM selects an appropriate number of steps}
\label{sec:discuss-a}
To better understand the behavior of SUTA-LM, we compare its average number of selected adaptation steps to that of an \textit{oracle} strategy.
The \textit{oracle} strategy selects the step with the lowest WER, breaking ties by choosing the earliest occurrence across $N$ adaptation steps.

\begin{figure}[t]
\centering
\includegraphics[width=1.0\linewidth]{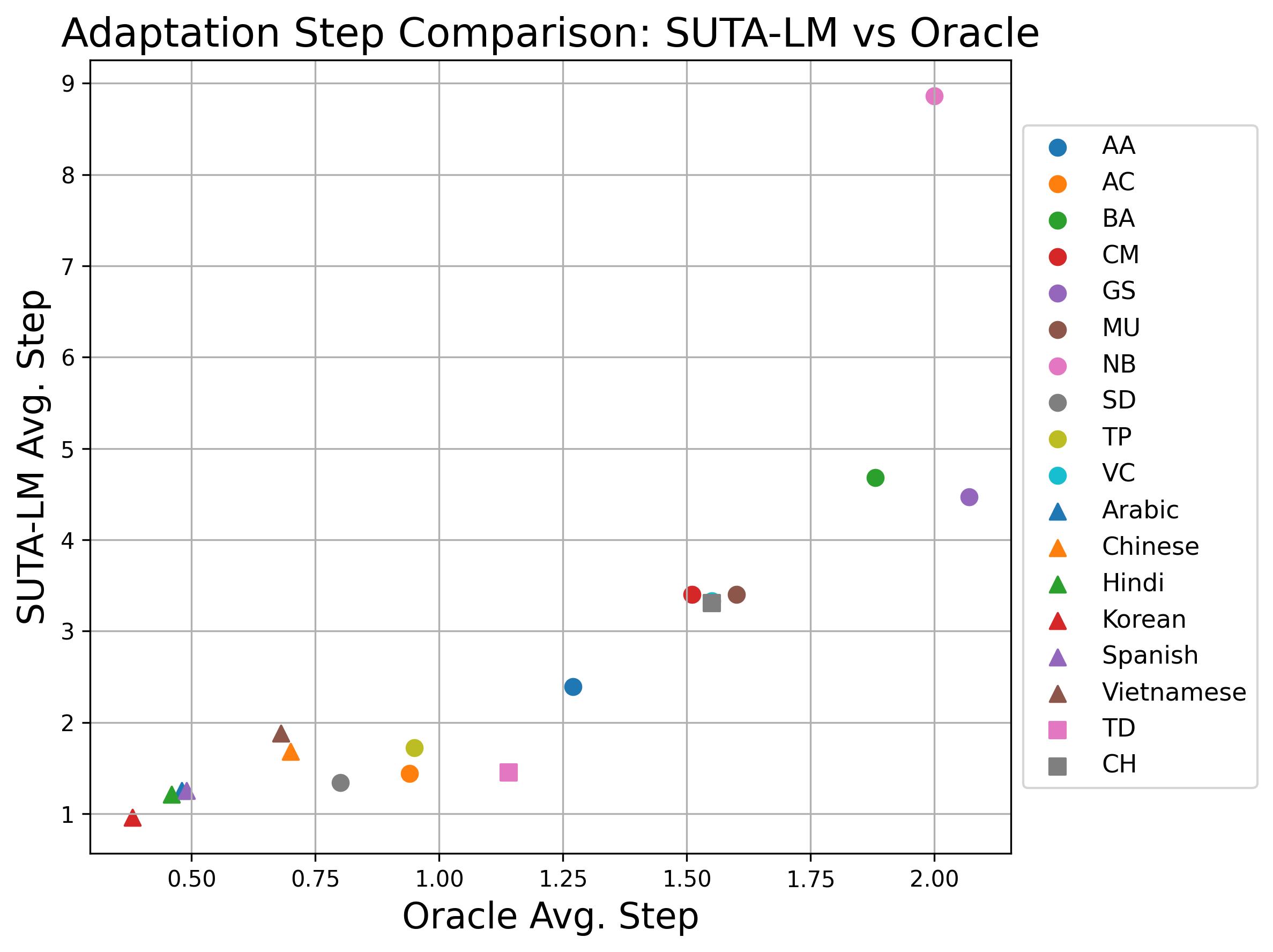}
\caption{Comparison of the average number of adaptation steps selected by SUTA-LM to the oracle strategy over 18 datasets.}
\label{fig:step-comp}
\end{figure}

As shown in Fig.~\ref{fig:step-comp}, SUTA-LM adaptively selects the number of adaptation steps based on the domain, and the trend closely matches that of the oracle. When fewer adaptation steps are sufficient, SUTA-LM uses fewer steps, and vice versa. These results suggest that SUTA-LM successfully adapts the number of steps to domain-specific characteristics in a dynamic and effective manner.

From another perspective, Table~\ref{tab:main-exp} shows that SUTA outperforms Rescoring in cases with larger acoustic shifts, such as CH and LS-C. In contrast, Rescoring performs better in domains like TD and L2Arctic, where adaptation provides less benefit. Correspondingly, SUTA-LM selects more adaptation steps on domains like BA, GS, and NB, where SUTA is stronger, and fewer on domains like Korean, Spanish, and Vietnamese, where Rescoring dominates. The average number of steps selected by SUTA-LM aligns well with the trend, further validating its ability to balance adaptation and rescoring based on input characteristics.

\begin{table*}[t]
  \centering
  \caption{Ablation on different components of SUTA-LM. We report results on a representative subset. Reported WER(\%) is averaged over 3 runs.}
      \begin{tabular}{lcccccc}
        \toprule
         \textbf{Method} & \textbf{Korean} & \textbf{Spanish} & \textbf{TD} & \textbf{CH} & \textbf{GS} & \textbf{NB}\\
        \midrule
        \textbf{SUTA-LM} & \textbf{11.7} & 15.0 & \textbf{11.0} & \textbf{22.4} & \textbf{27.0} & \textbf{63.8} \\
        \midrule
        \textbf{w/o Acoustic Score Thresholding} & \textbf{11.7} & \textbf{14.8} & \textbf{11.0} & 22.9 & 27.5 & 64.0 \\
        \textbf{w/o Early Stop}\\
        \textit{\quad Acoustic Score Thresholding + Linguistic Score Selection} & 11.8 & 15.1 & 11.1 & 22.5 & \textbf{27.0} & 63.9 \\
        \textit{\quad Acoustic Score Thresholding + Random Selection} & 13.2 & 16.3 & 11.3 & 22.8 & 27.4 & 64.2 \\
        \bottomrule
      \end{tabular}
  \label{tab:ablate}
\end{table*}

\subsection{Comparison with Fixed-Step Adaptation}
\label{sec:discuss-b}
Using a fixed number of adaptation steps is common in previous works on TTA for ASR. However, as discussed in the preliminary study, this approach is suboptimal, as different datasets might require different numbers of adaptation steps.

As shown in Fig.~\ref{fig:prelim2}, SUTA-LM performs comparably to the best possible fixed-step selection strategy, requiring no prior information. These findings suggest that future TTA methods should avoid relying on fixed-step adaptation and instead adopt more flexible, input-aware strategies.

\subsection{Ablation Study}
\label{sec:ablation}
Table~\ref{tab:ablate} summarizes the results of the ablation study on different components of SUTA-LM.  Due to space constraints, we report results for a representative subset of six datasets: Korean, Spanish, TD, CH, GS, and NB.

First, removing acoustic score thresholding leads to clear performance degradation on domains with larger acoustic mismatch, such as CH, GS, and NB. In these cases, relying solely on linguistic scores often causes the model to stop adaptation prematurely, selecting steps where the ASR model is still underconfident, even though the linguistic score $p_{\text{lm}}(y_t)$ appears high. On the contrary case, ASR is already confident even in the early stage; therefore, removing acoustic score thresholding has less impact on these domains.

Second, comparing SUTA-LM with and without early stopping shows nearly identical performance, indicating that the auto-step mechanism effectively avoids unnecessary adaptation. This results in notable computational savings shown in Section~\ref{sec:results} but without any loss in accuracy.

Finally, we evaluate the importance of linguistic score selection by replacing it with a random step selection baseline, both carried out without early stopping. As shown in Table~\ref{tab:ablate}, random selection significantly degrades performance, especially on Korean (WER increases from 11.8 to 13.2) and Spanish (WER increases from 15.1 to 16.3). This effect is more pronounced in domains with less acoustic mismatch, where a broader range of adaptation steps may appear acoustically reasonable, i.e., with an acoustic score larger than the threshold $\tau$. In such cases, linguistic information plays a critical role in filtering out overconfident yet erroneous transcriptions, highlighting the value of linguistic score selection.

Overall, these results confirm that all components of SUTA-LM contribute meaningfully to its performance. By dynamically selecting the appropriate adaptation step for each input, SUTA-LM achieves robust performance across diverse conditions while remaining computationally lightweight.

\subsection{Different Acoustic Threshold}
We explore different acoustic thresholds for SUTA-LM on the representation subset described in Section~\ref{sec:ablation}. Table~\ref{tab:ablate3} reports the results for various values of $\tau \in \{-0.01, -0.03, -0.05, -0.1, -0.3\}$. While extreme values lead to suboptimal performance, a moderate threshold yields robust results across domains. Our setting balances between filtering out unreliable steps and retaining potential candidates.

\begin{table}
  \centering
  \caption{Averaged WER(\%) comparison of different acoustic thresholds on a representative subset.}
    \begin{tabular}{c|ccccc}
        \toprule
        $\tau$ & -0.01 & -0.03 & -0.05 & -0.1 & -0.3\\
        \midrule
        Avg. WER & 26.2 & 25.6 & \textbf{25.2} & \textbf{25.2} & 25.3\\
        \bottomrule
    \end{tabular}
  \label{tab:ablate2}
\end{table}

\subsection{Different Source ASR Models}
\label{sec:model-ablation}
To test the generalization of the proposed method, we adopt other source ASR models with SUTA-LM. Following \cite{lin-etal-2024-continual}, besides wav2vec 2.0-base, we test on data2vec-base\footnote{https://huggingface.co/facebook/data2vec-audio-base960h}, and HuBERT-large\footnote{https://huggingface.co/facebook/hubert-large-ls960-ft} model. All the ASR models are trained with Librispeech 960 hours. We report average results on the representative subset described in Section~\ref{sec:ablation}, as shown in Table~\ref{tab:ablate2}. SUTA-LM performs effectively across different models, demonstrating its broad applicability.

\begin{table}
  \centering
  \caption{Averaged WER(\%) comparison of different CTC-based ASR models on a representative subset.}
    \begin{tabular}{lccc}
        \toprule
        \textbf{Method} & wav2vec2-base & data2vec-base & hubert-large\\ \midrule
        \textbf{Source Model} & 34.8 & 30.9 & 19.4\\
        \midrule
        \textbf{SUTA} & 26.6 & 26.1 & 16.9\\
        \textbf{Rescoring} & 28.3 & 25.3 & 16.4\\
        \textbf{SGEM} & 26.3 & 26.2 & 17.1\\
        \textbf{SUTA+Rescoring} & 26.8 & 24.9 & \textbf{16.2}\\
        \textbf{SUTA-LM} & \textbf{25.2} & \textbf{24.7} & \textbf{16.2}\\
        \bottomrule
    \end{tabular}
  
  \label{tab:ablate3}
\end{table}

\section{Conclusion}
In this work, we propose a novel approach to effectively combine TTA and LM rescoring by introducing an auto-step selection mechanism. Through a preliminary analysis, we explore the interplay between TTA and LM rescoring in ASR, revealing that inappropriate adaptation can interfere with LM rescoring and highlighting the need for adaptive control. Building on this insight, we present SUTA-LM, an extension of SUTA with LM rescoring that dynamically selects the appropriate adaptation step for each input. Extensive evaluations across diverse domains show that SUTA-LM delivers robust, consistent performance with high efficiency. Our findings underscore the importance of carefully integrating TTA and LM rescoring and motivate future work on adaptive, input-aware strategies in this space.

\bibliographystyle{IEEEtran}
\bibliography{ref}

\end{document}